\documentclass[journal]{IEEEtran}
 \usepackage{graphicx}% more modern

\newif\ifReview
\newif\ifCitetActive
\newif\ifWithSectionNumber
\WithSectionNumbertrue

\newif\ifWithExtrinsicEvaluation
\newif\ifOmitAppendix
\OmitAppendixtrue

\newif\ifSuppressMemo
\ifSuppressMemo
\newcommand{\memo}[1]{}
\else
\usepackage{color}
\newcommand{\memo}[1]{{\bf \color{red}{#1}} \color{black}}
\fi

\usepackage[utf8]{inputenc} % allow utf-8 input
\usepackage[T1]{fontenc}    % use 8-bit T1 fonts
\usepackage{hyperref}       % hyperlinks
\usepackage{url}            % simple URL typesetting
\usepackage{booktabs}       % professional-quality tables
\usepackage{amsfonts}       % blackboard math symbols
\usepackage{nicefrac}       % compact symbols for 1/2, etc.
\usepackage{microtype}      % microtypography

\usepackage{url}
 \usepackage{epstopdf}
\usepackage{subfigure} 
\usepackage{amssymb}
\usepackage{amsmath}
\usepackage{amsbsy}
\usepackage{latexsym}
\usepackage{multirow}

\usepackage{enumitem}
\usepackage[normalem]{ulem}
\usepackage{booktabs}
\usepackage{microtype}

\usepackage{algorithm}
\usepackage{algpseudocode,algorithmicx}

\usepackage{nkj,nkj_e}

\newcommand{\citep}{\cite}
\newcommand{\citet}{\cite}

\ifCitetActive
\newcommand{\citett}[2]{\citet{#2}}
\else
\newcommand{\citett}[2]{#1 \cite{#2}}
\fi

\ifCLASSINFOpdf
\else
\fi
\begin{document}
\title{SynsetRank: Degree-adjusted Random Walk \\ for Relation Identification}

\author{
Shinichi Nakajima$^{*}$, Sebastian Krause, Dirk Weissenborn, Sven Schmeier, Nico G{\"o}rnitz, Feiyu Xu
%Michael~Shell,~\IEEEmembership{Member,~IEEE,}
%        John~Doe,~\IEEEmembership{Fellow,~OSA,}
%        and~Jane~Doe,~\IEEEmembership{Life~Fellow,~IEEE}% <-this % stops a space
\thanks{$^*$corresponding author (email: nakajima@tu-berlin.de)}
\thanks{Shinichi Nakajima and Nico G{\"o}rnitz are with Technische Unversit{\"a}t Berlin, Machine Learning Group, Marchstr. 23, 10587 Berlin, Germany.}% <-this % stops a space
\thanks{Shinichi Nakajima, Sebastian Krause, Dirk Weissenborn, Sven Schmeier and Feiyu Xu are with Berlin Big Data Center, 10587 Berlin, Germany.}% <-this % stops a space
\thanks{Sebastian Krause, Dirk Weissenborn, Sven Schmeier and Feiyu Xu are with DFKI, Language Technology Lab, Alt-Moabit 91c, Berlin, Germany.}% <-this % stops a space
}


\markboth{Journal of \LaTeX\ Class Files,~Vol.~14, No.~8, August~2015}%
{Nakajima \MakeLowercase{\textit{et al.}}: SynsetRank: Degree-adjusted Random Walk \\ for Relation Identification}



\maketitle

\begin{abstract}
In relation extraction,
a key process is to obtain good \emph{detectors}
that find relevant sentences describing the target relation.
To minimize the necessity of labeled data for refining detectors,
previous work successfully made use of BabelNet, a semantic graph structure expressing 
relationships between \emph{synsets},
as side information or prior knowledge.
The goal of this paper is to enhance the use of graph structure
in the framework of random walk with a few adjustable parameters.
Actually, a straightforward application of random walk degrades the performance even after parameter optimization.
With the insight from this unsuccessful trial, we propose
\emph{SynsetRank}, which adjusts the initial probability so that high degree nodes
influence the neighbors as strong as low degree nodes.
In our experiment 
on 13 relations in 
the \texttt{FB15K-237} dataset,
SynsetRank 
significantly outperforms baselines and the plain random walk approach.
\end{abstract}

\begin{IEEEkeywords}
relation extraction, random walk, PageRank, BabelNet.
\end{IEEEkeywords}

% \fi
\IEEEpeerreviewmaketitle

\section{Introduction}
%SK: copy a few sentences from iswc 2013 paper, write short motivating blurb
%
Many NLP tasks are concerned with recognizing semantic concepts in large amounts of text, including the problem of detecting mentions of real-world events \cite{Alfonseca2013,ZhangSW2015} and relations between entities \cite{ZhouSZZ05,NakasholeWS12,MitchellCHTBCMG15}.
The high up-front cost for training NLP systems with labeled data has lead to the design of supervision paradigms which use only \emph{distant} or \emph{weak} guidance from manually created examples \cite{Zhang2004,MintzBSJ2009,HoffmannZLZW2011}. Such methods can benefit from additional clues about the presence of semantic concepts in language fragments, coming from lexical-semantic resources like WordNet \cite{Fellbaum1998} and BabelNet \cite{NavigliP2012}.

Consider the following example:
%
%\begin{example}
% \textit{\uline{Mike and Julie Miller} exchanged rings last summer, only two years after \uline{they} had first met.}
\textit{\uline{Mike and Julie Miller} celebrated a fabulous wedding last summer, only two years after \uline{they} had first met.}
%\end{example}
%
In the first part of the sentence, the term \textit{wedding} is a strong indicator for the presence of a \textit{marriage} relation mention, while the second part has no such indicator. Information about such relation-relevant terms is helpful for, e.g., extracting sentence templates for pattern-based relation extraction, or pre-filtering texts before fine-grained processing takes place.

Existing information-extraction systems typically exploit lexical-semantic repositories for increased lexical coverage, by retrieving synonyms for observed terms or by calculating similarity scores based on the graph structure of these resources \cite{CulottaS04,Stevenson2005,ZhouZ07}. Few approaches explicitly identify the entries which express semantic concepts on the textual level. An exception is the work by \citett{Moro et al. (2013)}{Moro13}, who start with an initial frequency distribution of terms co-occurring with relation examples in a large text collection. Relation-revelant terms are then determined through an ad-hoc combination of this initial distribution with the graph structure of the repository.

In this paper, we improve Moro et al.'s approach by casting the problem to a ranking problem, and applying the random walk approach with a simple modification. We test our approach on a publicly available dataset and compare it to several baselines. 
\ifWithExtrinsicEvaluation
In contrast to \cite{Moro13}, we directly evaluate the model performance in terms of the quality of positively labeled word synsets and reach drastically better performance. Testing the impact of this method on the quality of downstream applications remains as future work for now.
\else
%In contrast to \cite{Moro13}, 
We evaluate the model performance in terms of the quality of positively labeled word synsets and reach drastically better performance. 
%Testing the impact of this method on the quality of downstream applications remains as future work for now.
\fi
\section{Background}

This paper focuses on the automatic identification of relation-relevant entries in lexical-semantic resources, i.e., we want to obtain relation \emph{detectors}. As \emph{relation} we understand any kind of real-world relationship between persons, locations, etc., examples are the \textit{kinship} relations \textit{marriage}, \textit{parent-child}, \textit{siblings}, or business concepts such as \textit{company acquisition}, \textit{employment tenure}.

\emph{Lexical-semantic repositories} are inventories of word senses, which link words to their meaning and to other words, i.e., these resources have an underlying graph structure. 
A prominent instance of the many lexical-semantic resources out there is BabelNet\footnote{\url{http://babelnet.org}} \cite{NavigliP2012}, a large-scale multilingual semantic network which was built automatically through the algorithmic integration of Wikipedia and WordNet.
The core components (nodes) are so-called \emph{synsets}, which are sets of synonymous terms; the edges correspond to synset relationships such as hypernymy and meronymy.

\subsection{Finding Domain-Relevant Terms}
A lot of work has dealt with acquiring relevant terms for semantic relations. \citett{Nguyen et al. (2010)}{Nguyen2010} analyzed the distribution of \emph{trigger words} for semantic relations in annotated data in order to filter extraction patterns. For a similar reason, \citett{Xu et al. (2002)}{XuKPS02} collected relevant terms with a TFIDF-based strategy. Other approaches incorporate lexical knowledge from WordNet. \citett{Zhou et al. (2005)}{ZhouSZZ05} presented a feature-based relation extractor which utilizes semi-automatically build trigger-word lists from WordNet. \citett{Culotta and Sorensen (2004)}{CulottaS04} used WordNet hypernyms for increased extraction coverage. \citett{Stevenson and Greenwood (2005)}{Stevenson2005} defined a similarity function for learned linguistic patterns that was built on WordNet information.

None of the above approaches, however, explicitly determines and outputs which parts of the lexical-semantic resource contain the terms that are relevant to a given semantic relation.

\subsection{Extracting Relation-Specific Sub-Graphs}
\label{sec:MorosMethod}
\citett{Moro et al. (2013)}{Moro13} proposed another approach to the term identification problem. Their algorithm gets as input a set of sentences which have been labeled with relation mentions in a distantly supervised manner. This noisy set of relation mentions is processed by word-sense disambiguation \cite{Navigli2009} to build links from the word level to the level of synsets in WordNet and BabelNet. This induces a frequency distribution over synsets, from which the most frequent items plus their direct neighbors in the resource are selected to build the final relation-specific sub-graph. Moro et al. (2013) employ these sub-graphs for filtering linguistic patterns in a relation-extraction scenario. While their approach shows good results, it also leaves room for improvements mainly due to its ad-hoc, heuristic utilization of the available synset links in the sense inventory. We use their approach as one of the baselines against which we compare our proposed model.

\subsection{PageRank: Random Walk for Webpage Ranking}

Moro et al. (2013) choose the most frequent synsets and their neighbors as the relevant synsets,
which can be naturally cast as information propagation through random walks.
%\citett{Moro et al. (2013)  resembles random walks,
%where a property of each node is propagated to its neighbors.
Random walk was
successfully applied for ranking webpages
in the name of \emph{PageRank} \cite{Brin98,Leskovec14}.
Our first trial is to apply PageRank for ranking synsets, 
according to the relevance to the target relation.

Consider a graph $\mcG = (\mcV, \mcE)$
with a set $\mcV$ of nodes and a set $\mcE$ of edges connecting
two nodes.  We denote the number of nodes by $N = |\mcV|$.
In our application, each node $i \in \mcV$ corresponds to a synset,
and each edge $(i, j) \in \mcE$ corresponds to a semantic connection between two synsets.
Each edge has a label $l \in \{1, \ldots, L\}$, which specifies the relation between two synsets, %, e.g., %synonymy, 
such as
hypernymy and meronymy.

We formally prepare $L$ graphs $\{G^{(l)} = (\mathcal{V}, \mathcal{E}^{(l)})\}_{l=1}^L$,
which share the nodes $\mcV$ but have $L$ different sets of edges,
each of which consists of a single edge label.
For each $l$, we express the existence of edges 
by an $N \times N$ binary matrix $\bfE^{(l)}$,
and we prepare a weight vector $\bfw \in \mathbb{R}_{+}^{L}$.
Then, we construct a transition matrix $\bfQ' \in \mathbb{R}^{N \times N}$ as the weighted sum over all edge labels:
\begin{align}
Q'_{i, j}
&= \textstyle
\begin{cases}
\textstyle
\frac{\sum_{l=1}^L  w_l E_{i, j}^{(l)}}{\sum_{j'=1}^N \sum_{l=1}^L  w_l E_{i, j'}^{(l)}} & \mbox{ if } (i, j) \in \bigcup_{l=1}^L \mcE^{(l)},\\
0 & \mbox{ otherwise}.
\end{cases}
\notag
%\label{eq:DefinitionQPrime}
\end{align}
The BabelNet graph is directed, and the number of edge types is $L=29$.
Since the semantic edge direction does not necessary indicates the direction
in which the relevance information should flow,
we treat the edges in the opposite direction as another edge type.
Thus, we have $L = 58$ edge types in total.

To avoid \emph{dead-ends} and \emph{spider-traps} issues,
our implementation of
Page\-Rank is equipped with \emph{taxation} and \emph{restarting} \cite{Leskovec14,Tong06}.
This can be realized by adding a \emph{sink-source} node,
which absorbs $\alpha \in [0, 1]$ proportion of flow from all nodes,
and re-distributes it according to the initial distribution ${\bfp}^{(0)} \in \mathbb{R}_+^{N}$.
Here, we use the original frequency distribution, observed from the text corpus
(see Section~\ref{sec:MorosMethod}),
 over synsets
as the initial distribution ${\bfp}^{(0)}$.

We also add self-links,
with which the random walkers stay at the same node with probability 
$\beta \in [0, 1]$.
Thus, the distribution after $t$ random walks is
defined as
%the equilibrium probability $\bfp \in \mathbb{R}_+^{N+1}$ satisfies
\begin{align}
\widetilde{\bfp}^{(t)\T} &= \widetilde{\bfp}^{(t-1)\T} \bfQ, \qquad \mbox{where}
%\notag\\
\label{eq:EquilibriumCondition}\\
\bfQ &= 
\textstyle
\begin{pmatrix}
(1 - \alpha) \{(1 - \beta)\bfQ' + \beta \bfI_N \}& \bfk\\
 {\bfp}^{(0)\T}& 0
\end{pmatrix},
\notag \\
k_{i} &= 
%\alpha^{\sum_{j} Q'_{i, j}}
%=
\begin{cases}
\alpha \qquad \mbox{ if } {}^\exists j' \mbox{ s.t. } (i, j') \in  \bigcup_{l=1}^L \mcE^{(l)},\\
1 - (1 - \alpha) \beta\qquad \mbox{ otherwise}.
\end{cases}
\notag
\end{align}
%for $\alpha \in (0, 1)$.
%Note that, for $Q'$ defined by Eq.\eqref{eq:DefinitionQPrime},
%$\sum_{j} Q_{i, j}' $ can be equal to zero or one.
Here, $\bfI_N$ denotes the $N \times N$ identity matrix,
and 
$\widetilde{\bfp}^{(t)} \in \mathbb{R}^{N+1}$ denotes 
the distribution (after $t$ random walks) over the original nodes and the $(N+1)$-th sink-source node.
We set $\widetilde{\bfp}^{(0)}$ to the initial distribution augmented with a zero for the sink-source node, i.e., $\widetilde{\bfp}^{(0)} = (\bfp^{(0)\T}, 0)^\T$.
After random walks,
we rank the synsets based on the distribution $\widetilde{\bfp}^{(t)}$.
The unknown parameters $\alpha, \beta$, and $t$ are optimized by using the validation data,
while the edge weights are fixed to $w_l = 1, \forall l$ in this paper.

\section{Proposed Method}

As shown in 
\ifWithSectionNumber
Section~\ref{sec:Experiment},
\else
the next section,
\fi
PageRank performs worse than Moro et al.'s baseline method %\newcite{Moro13}
even after parameter optimization.
This is not very surprising because the problem of ranking webpages
and the problem of ranking synsets are substantially different.
Taking this difference into account,
we propose a new method.

\subsection{SynsetRank: Degree-Adjusted Random Walk for Synset Ranking}

By nature of random walks,
a node with more outgoing edges 
influences each neighboring node less,
%less influences each neighboring node,
since random walkers are dispersed over many edges.
This is what PageRank, which simulates web surfers, 
intends to do, but this is not appropriate for synset ranking,
where neighbors to frequent synsets should be 
high ranked, regardless of the degree (the number of edges) 
of the frequent node.

Our idea is to adjust the original frequency, as well as the restarting probability,
to compensate this undesired phenomenon.
In our random walk formulation \eqref{eq:EquilibriumCondition},
this can be done simply by replacing the original frequency distribution $\bfp^{(0)}$ with
a re-weighted one:
% \begin{align}
% \widehat{\bfp}^{(0)\T} & \!\!=\!  \bfp^{(0)\T} \! \bfdiag(\bfd),  \mbox{ where } 
% d_i \!=\! \textstyle \frac{\sum_{j=1}^N Q'_{i, j}}
%       {\sum_{i'=1}^N  \sum_{j=1}^N Q'_{i', j}}.
% \notag
% \end{align}
% \begin{align}
% \widehat{\bfp}^{(0)\T} & \!\!=\!  \frac{\bfp^{(0)\T} \! \bfdiag(\bfd)}{\|\bfp^{(0)\T} \! \bfdiag(\bfd)\|_1},  \mbox{ where } 
% d_i \!=\! \textstyle \sum_{j=1}^N \sum_{l=1}^L  w_l E_{i, j}^{(l)}.
% \notag
% \end{align}
% Here $\bfdiag(\bfd)$ denotes the diagonal matrix with the diagonal elements specified by $\bfd = (d_1, \ldots, d_N)$.
\begin{align}
\widehat{\bfp}^{(0)} & \!\!=\! \textstyle \frac{\bfp^{(0)} * \bfd}{\|\bfp^{(0)} * \bfd\|_1},  \mbox{ where } 
d_i \!=\! \textstyle \sum_{j=1}^N \sum_{l=1}^L  w_l E_{i, j}^{(l)}.
\notag
\end{align}
Here $*$ denotes the element-wise product of vectors.
This simple modification makes the influence of a node to each neighbor equal, regardless of the degree, and shows a drastic improvement in our experiment in 
\ifWithSectionNumber
Section~\ref{sec:Experiment}.
\else
the next section.
\fi
We call this degree-adjusted random walk approach \emph{SynsetRank}.

\if 0

\else

\def\dummyinterval{1.7mm}

\begin{table*}[t]
\centering
\caption{Area under the ROC curve (AUC) for 13 relations of the \texttt{FB15k-237} dataset.}
\label{table:Result}
\small
\resizebox{\textwidth}{!}{%
\begin{tabular}{@{}lcccccc@{}}
\toprule
Relation & Frequency & Moro & PageRank & PageRank &  SynRank &  SynRank 
\\        &        &     &         & (common) &          &   (common)  \\
\midrule
 \shortstack[l]{\scriptsize\itshape /award/award\_nominee/award\_nominations\allowbreak \\ \;\;\; \scriptsize\itshape./award/award\_nomination/award\_nominee}
% \footnotesize\itshape /award/award\_nominee/award\_nominations\ldots
&    0.4120  &  0.4666  &  0.4698  &  0.3794  &  0.4398  & {\bf 0.4868}   \\
 \shortstack[l]{\scriptsize\itshape /award/award\_winner/awards\_won\allowbreak \\ \;\;\; \scriptsize\itshape ./award/award\_honor/award\_winner}
% \footnotesize\itshape /award/award\_winner/awards\_won\ldots
&   {\bf 0.6204}  &  0.4625  &  0.5079  &  0.5378  &  0.4847  &  0.5022   \\
 \shortstack[l]{\scriptsize\itshape /base/popstra/celebrity/friendship\allowbreak \\ \;\;\; \scriptsize\itshape./base/popstra/friendship/participant}
% \footnotesize\itshape /base/popstra/celebrity/friendship\ldots
&    0.3825  &  0.4162  &  0.4216  &  0.4084  &  {\bf0.4824}  &  0.4232   \\
 \shortstack[l]{\scriptsize\itshape /education/educational\_degree/people\_with\_this\_degree\allowbreak \\ \;\;\; \scriptsize\itshape./education/education/institution}
% \footnotesize\itshape /education/educational\_degree/people\_with\_this\_degree\ldots
&    0.7002  &  0.8573  &  0.7128  &  0.6355  &  {\bf0.8743}  &  0.8646   \\
 \shortstack[l]{\scriptsize\itshape /education/educational\_institution/students\_graduates \allowbreak \\ \;\;\; \scriptsize\itshape./education/education/major\_field\_of\_study}
% \footnotesize\itshape /education/educational\_institution/students\_graduates\ldots
&    0.6803  &  0.7710  &  0.7099  &  0.7239  &  0.7726  &  {\bf0.7891}   \\
 \shortstack[l]{\scriptsize\itshape /film/actor/film./film/performance/film \allowbreak \\ \;\;\; \vspace{\dummyinterval} }
% \footnotesize\itshape /film/actor/film./film/performance/film
&    0.7149  &  0.6519  & {\bf 0.7228}  &  0.6998  &  0.6694  &  0.6479   \\
 \shortstack[l]{\scriptsize\itshape /film/director/film \allowbreak \\ \;\;\; \vspace{\dummyinterval} }
% \footnotesize\itshape /film/director/film
&    0.6237  &  0.7116  &  0.5760  &  0.5762  & {\bf 0.6873 } &  0.6847   \\
 \shortstack[l]{\scriptsize\itshape /location/location/contains \allowbreak \\ \;\;\; \vspace{\dummyinterval} } 
% \footnotesize\itshape /location/location/contains
&    0.5856  & {\bf 0.6747}  &  0.5858  &  0.5682  &  0.6624  &  0.6709   \\
\shortstack[l]{\scriptsize\itshape /music/performance\_role/regular\_performances\allowbreak \\ \;\;\; \scriptsize\itshape./music/group\_membership/role}
% \footnotesize\itshape /music/performance\_role/regular\_performances\ldots
&    0.6373  &  0.7483  &  0.6717  &  0.6666  &  {\bf0.8185}  &  0.8137   \\
 \shortstack[l]{\scriptsize\itshape /organization/organization\_member/member\_of\allowbreak \\ \;\;\; \scriptsize\itshape./organization/organization\_membership/organization}
% \footnotesize\itshape /organization/organization\_member/member\_of\ldots
&  {\bf  0.9200}  &  0.8020  &  0.9126  &  0.8876  & {\bf 0.8812}  &  0.8230   \\
 \shortstack[l]{\scriptsize\itshape /people/person/nationality \allowbreak \\ \;\;\; \vspace{\dummyinterval} }
% \footnotesize\itshape /people/person/nationality
&    0.5507  &  0.7180  &  0.8411  &  0.7413  &  {\bf0.8820}  &  0.8399   \\
 \shortstack[l]{\scriptsize\itshape /people/person/place\_of\_birth \allowbreak \\ \;\;\; \vspace{\dummyinterval} }
% \footnotesize\itshape /people/person/place\_of\_birth
&    0.5905  &  0.6977  &  0.7792  &  0.6914  & {\bf 0.8315}  &  0.7796   \\
 \shortstack[l]{\scriptsize\itshape /people/person/places\_lived\allowbreak \\ \;\;\; \scriptsize\itshape./people/place\_lived/location}
% \footnotesize\itshape /people/person/places\_lived\ldots
&    0.6573  &  0.7203  &  0.7534  &  0.7357  &  {\bf0.8238}  &  0.7861   \\
\midrule
Average AUC &    0.6212  &  0.6691  &   0.6665  &  0.6348  & {\bf 0.7162}  &  0.7009   \\
\bottomrule
\end{tabular}%
} % END OF resizebox{}
\end{table*}

\fi

\section{Experiment}
\label{sec:Experiment}

In this section, we show our experimental results.

\subsection{Data and Task}
We used the \texttt{FB15K-237} dataset of \citett{Toutanova et al. (2015)}{Toutanova2015EMNLP} for our experiments, and follow the training/validation/test split suggested by the authors. This dataset provides a large number of relation instances from the factual knowledgebase Freebase along with textual mentions, i.e., parses of sentences which contain references to the argument entities of the relation instances. 

The task is to find the synsets (nodes in BabelNet) that are semantically relevant for the target relation, i.e., its occurrence is likely to trigger a specific semantic relation (from a knowledge base like Freebase, e.g., the relation {\itshape/people/person/place\_of\_birth} connecting humans to the place they were born in). Here, `triggering' means that this synset (word surface form) in a sentence (e.g., the word `born' in the sentence `John was born in New York'), makes it probable that this sentence refers to this semantic relation (which the sentence actually does, in this example). 
Such information about relation relevancy is useful for downstream text analytics tasks where it serves as a further signal for making a relation extraction decision (Does the sentence `John was born in New York' contain the fact triple <John, {\itshape/people/person/place\_of\_birth}, New York>?). 

The dataset \texttt{FB15K-237}  was created by combining (a) fact triples from Freebase, (b) many sentences which mention entities for which facts are listed in Freebase. 
%We follow the general idea that lexical-semantic resources (=BabelNet) can help in making good extraction decisions, and propose a specific method in this direction. 
As the task for which \texttt{FB15K-237} was created, unfortunately, is different from ours,
we cannot follow the evaluation procedure suggested in \citett{Toutanova et al. (2015)}{Toutanova2015EMNLP}.
Accordingly, we created our own gold-standard labels by hand-labeling a subset of synsets,
as explained shortly.%
\footnote{We will make the evaluation data publicly available upon acceptance.
%, see supplementary \texttt{paper-data.tgz}.
}

In order to avoid data sparsity issues, we determined the twenty relations with the highest number of mentions in the training partition, and removed seven from these which were redundant with respect to the other relations or which were semantically lightweight from the point of view of textual mentions
(see Appendix~\ref{sec:RemovedRelations} for details of the removed seven relations).
%
%\footnote{For example, we excluded a relation dealing with movie release dates per world region.
%Appendix~\ref{sec:RemovedRelations} details the removed seven relations.
%}
%\footnote{\url{/award/award_nominee/award_nominations./award/award_nomination/nominated_for
%/award/award_winning_work/awards_won./award/award_honor/award_winner
%/location/location/adjoin_s./location/adjoining_relationship/adjoins
%/location/hud_county_place/place
%/film/film/release_date_s./film/film_regional_release_date/film_release_region
%/film/film/country
%/influence/influence_node/influenced_by}
%
%The first two are semantically very close to the first used relation (see Table1), the third and fourth are to the used 8th,
%and the fifth and sixth are to the used 7th. 
%`Semantically close' means that the synsets relevant for the respective relations are very likely to be the same,
%and we can expect that the results on them are similar to the close used relations.
%The seventh discarded relation is unlikely to be mentioned at all in the text, so is not interesting for relation extraction.
%We discarded these relations to minimize the labelling work (8 person-hours per relation) by our colleagues.}
We used the 13 relations shown in the first column of Table~\ref{table:Result}.
%We used the following 13 relations in our experiments:
% \begin{itemize}[noitemsep,topsep=0pt,parsep=0pt,partopsep=0pt]
% \scriptsize\itshape
% \item /location/location/contains
% \item /film/actor/film\allowbreak./film/performance/film
% \item /award/award\_nominee/award\_nominations\allowbreak./award/award\_nomination/award\_nominee
% \item /people/person/nationality
% \item /award/award\_winner/awards\_won\allowbreak./award/award\_honor/award\_winner
% \item /education/educational\_degree/people\_with\_this\_degree\allowbreak./education/education/institution
% \item /people/person/places\_lived\allowbreak./people/place\_lived/location
% \item /education/educational\_institution/students\_graduates\allowbreak./education/education/major\_field\_of\_study
% \item /organization/organization\_member/member\_of\allowbreak./organization/organization\_membership/organization
% \item /people/person/place\_of\_birth
% \item /film/director/film
% \item /base/popstra/celebrity/friendship\allowbreak./base/popstra/friendship/participant
% \item /music/performance\_role/regular\_performances\allowbreak./music/group\_membership/role
% \end{itemize}

For each of these relations and each data partition, we build positive/negative sets of textual mentions. The positive mentions are simply the ones that contain the arguments of a relation instance, while the negative ones are constructed following the strategy outlined by \citett{Toutanova et al. (2015)}{Toutanova2015EMNLP}. For the data in the training partition, we apply word-sense disambiguation to the positive and negative textual mentions, this way creating an initial synset frequency distribution among positive and negative examples for each relation, similar to \citett{Moro et al. (2013)}{Moro13}.

Furthermore, for each relation and the mentions in the validation and test partition, we prepare an evaluation dataset with manually annotated labels. We start with the top-50 most frequent synsets for the relation in the respective part of the data, which occur in positive textual mentions but not in negative ones. We do a two-step graph walk on BabelNet which extends these nodes with two randomly selected neighbors for each already included node. The resulting synsets and the corresponding lemmas/words are given to human annotators who label the synsets as positive/negative with respect to the relation, i.e., they judge whether or not the synset is relevant for the semantics of the relation.
% skrause@lns-87247:/local/tmp/skrause$ cat babelnet_graph.tsv | cut -f1 > tmp1 && cat babelnet_graph.tsv | cut -f3 > tmp2 && pv tmp1 tmp2 > tmp && sort -S 80% -u --parallel=30 -T /local/tmp/skrause tmp > tmp_sorted && wc -l tmp_sorted
%9106263 tmp_sorted
% skrause@lns-87247:/local/tmp/skrause$ cat babelnet_lemmas.tsv | cut -f2- | tr \\t \\n | wc -l
% 11029757
% skrause@lns-87247:/local/tmp/skrause$ pv babelnet_graph.tsv | wc -l
% 262676154
%% The sense inventory on which we test our approach is BabelNet\footnote{\url{http://babelnet.org}} \cite{NavigliP2012}, a large-scale multilingual semantic network which was built automatically through the algorithmic integration of Wikipedia and WordNet
In our experiments, we use BabelNet version 2.5.1, which contains roughly 9M synsets, 11M lexicalizations, and 262M links.
% sekr01@berlin-157:~/Dropbox/work_projects/201509-tub-svm-patterns/9th-iteration/FB15K-237-selected-relations/train$ cat -- */pos.txt | cut -f4 | tr -d \\r | gpaste -d + -s - | bc
%430444
% sekr01@berlin-157:~/Dropbox/work_projects/201509-tub-svm-patterns/11th-iteration/package-2016-03-29/test$ cat -- */intrinsic-dataset-positive.txt | wc -l
%    1027
%sekr01@berlin-157:~/Dropbox/work_projects/201509-tub-svm-patterns/11th-iteration/package-2016-03-29/test$ cat -- */intrinsic-dataset-negative.txt | wc -l
%   36118
There are in total 430k positive textual mentions for the 13 relations in the training partition of the data. The evaluation set has on average 2,857 synsets per relation, with a +/- ratio of 1:35.

%--------------------------------------
\subsection{Result}
%--------------------------------------
Table~\ref{table:Result} shows the area under the ROC curve (AUC) 
on the test partition for the 13 relations.
`Frequency' denotes the baseline method where the synsets are ranked 
based on the original frequency $\bfp^{(0)}$.
We clearly observe that the plain PageRank tends to perform worse than 
Moro et al.'s baseline method, while our proposed degree-adjusted SynsetRank
shows drastically better performance than all others.
For PageRank and SynsetRank, 
the optimal parameters (maximizing the AUC on the validation partition)
are grid-searched over $\alpha = 0.0, 0.2, \ldots, 1.0$,
$\beta = 0.0, 0.2, \ldots, 1.0$, and $t = 1, \ldots, 5$ for each relation.
For PageRank (common) and SynsetRank (common), the common optimal parameters
(maximizing the average AUC
over all 13 relations)
are used.

SynsetRank improves the average AUC of Moro et al.'s baseline method by roughly 0.05,
which is similar to the performance gain by \citett{Moro et al. (2013)}{Moro13} from the Frequency baseline.
Although the necessity of parameter optimization can be a bottleneck of SynsetRank,
the second best result with SynsetRank (common) implies the possibility of using common parameters
for all relations---Once we optimize the parameters for some set of relations,
one could use the same parameters for new relations.

%--------------------------------------
\section{Conclusion}
%--------------------------------------
Extracting knowledge from the internet is one of the most important near-future goals 
for researchers in the field of natural language processing, machine learning, and artificial intelligence.
Relation extraction (RE) is a key technology.
We cast the problem of finding good \emph{detectors} as a synset ranking problem,
and applied the random walk approach with simple modification.
Our experiment showed promising results.

We leave the quality assessment of downstream applications as future work.
We also plan to apply the \emph{supervised random walk} approach
\cite{Backstrom11} to optimize the weights $\bfw$ for each edge label, 
which further exploits existing knowledge for better performance.
% in addition to the other parameters, which are grid-searched in this paper.

\ifWithExtrinsicEvaluation
\appendices
\else
%\appendix
\appendices
\fi

\section{Removed seven relations}
\label{sec:RemovedRelations}

\begin{table}[t]
\caption{Removed seven relations.}
\label{table:OmittedRelations}
\centering
\small
%\resizebox{\textwidth}{!}{%
\begin{tabular}{lc|}
\toprule

 \shortstack[l]{\scriptsize\itshape /award/award\_nominee/award\_nominations\allowbreak \\ \;\;\; \scriptsize\itshape./award/award\_nomination/nominated\_for}
% \footnotesize\itshape /award/award\_nominee/award\_nominations \ldots 
\\
 \shortstack[l]{\scriptsize\itshape/award/award\_winning\_work/awards\_won\allowbreak \\ \;\;\; \scriptsize\itshape./award/award\_honor/award\_winner}
% \footnotesize\itshape /award/award\_winning\_work/awards\_won\ldots 
 \\
 \shortstack[l]{\scriptsize\itshape/location/location/adjoin\_s\allowbreak \\ \;\;\; \scriptsize\itshape./location/adjoining\_relationship/adjoins}
% \footnotesize\itshape /location/location/adjoin\_s\ldots 
\\
 \shortstack[l]{\scriptsize\itshape/location/hud\_county\_place/place\allowbreak \\ \;\;\; \vspace{\dummyinterval} }
% \footnotesize\itshape /location/hud\_county\_place/place
\\
 \shortstack[l]{\scriptsize\itshape/film/film/release\_date\_s\allowbreak \\ \;\;\; \scriptsize\itshape./film/film\_regional\_release\_date/film\_release\_region}
% \footnotesize\itshape /film/film/release\_date\_s\ldots 
\\
 \shortstack[l]{\scriptsize\itshape/film/film/country\allowbreak \\ \;\;\; \vspace{\dummyinterval} }
% \footnotesize\itshape /film/film/country
\\
 \shortstack[l]{\scriptsize\itshape/influence/influence\_node/influenced\_by\allowbreak \\ \;\;\; \vspace{\dummyinterval} }
% \footnotesize\itshape /influence/influence\_node/influenced\_by
 \\
\bottomrule
\end{tabular}
%} % END OF resizebox{}
\end{table}

Table~\ref{table:OmittedRelations} summarizes the removed seven relations in our experiment.
The first and the second \emph{removed} relations are semantically very close to the first and the second \emph{evaluated} relations,
respectively, in Table~\ref{table:Result}.
Likewise, the third and the fourth removed relations are close to the eighth evaluated relation,
and the fifth and the sixth removed relations are close to the seventh evaluated relation. 
Here, `semantically close' means that the synsets relevant for the respective relations are very likely to be the same,
and we can expect that the result on each removed relation is similar to its closest evaluated relation.
The seventh removed relation is unlikely to be mentioned at all in the text, so is not interesting for relation extraction.
Accordingly,
we removed those seven relations to reduce the hand-labeling work ($\sim8$ person-hours per relation).

\ifReview
\else
\section*{Acknowledgment}
%The authors thank the reviewers for helpful comments.
%Shinichi Nakajima thanks the support from 
SN, SK, DW, FX, SC thank the support from
%the German Research Foundation (GRK 1589/1)
%by the Federal Ministry of Education and Research (BMBF) 
%under the project Berlin Big Data Center (FKZ 01IS14013A).
the Berlin Big Data Center project (FKZ 01IS14013A).
NG was supported by ALICE II grant (FKZ 01IB15001B).

\fi

\ifCLASSOPTIONcaptionsoff
  \newpage
\fi



%\bibliographystyle{IEEEtran}

%\small
\bibliography{MachineLearning}

\begin{thebibliography}{10}
\providecommand{\url}[1]{#1}
\csname url@samestyle\endcsname
\providecommand{\newblock}{\relax}
\providecommand{\bibinfo}[2]{#2}
\providecommand{\BIBentrySTDinterwordspacing}{\spaceskip=0pt\relax}
\providecommand{\BIBentryALTinterwordstretchfactor}{4}
\providecommand{\BIBentryALTinterwordspacing}{\spaceskip=\fontdimen2\font plus
\BIBentryALTinterwordstretchfactor\fontdimen3\font minus
  \fontdimen4\font\relax}
\providecommand{\BIBforeignlanguage}[2]{{%
\expandafter\ifx\csname l@#1\endcsname\relax
\typeout{** WARNING: IEEEtran.bst: No hyphenation pattern has been}%
\typeout{** loaded for the language `#1'. Using the pattern for}%
\typeout{** the default language instead.}%
\else
\language=\csname l@#1\endcsname
\fi
#2}}
\providecommand{\BIBdecl}{\relax}
\BIBdecl

\bibitem{Alfonseca2013}
E.~Alfonseca, D.~Pighin, and G.~Garrido, ``{HEADY}: {N}ews headline abstraction
  through event pattern clustering.'' in \emph{ACL}, 2013, pp. 1243--1253.

\bibitem{ZhangSW2015}
C.~Zhang, S.~Soderland, and D.~S. Weld, ``Exploiting parallel news streams for
  unsupervised event extraction,'' \emph{{TACL}}, vol.~3, pp. 117--129, 2015.

\bibitem{ZhouSZZ05}
G.~Zhou, J.~Su, J.~Zhang, and M.~Zhang, ``Exploring various knowledge in
  relation extraction,'' in \emph{{ACL}}.\hskip 1em plus 0.5em minus
  0.4em\relax The Association for Computer Linguistics, 2005.

\bibitem{NakasholeWS12}
N.~Nakashole, G.~Weikum, and F.~M. Suchanek, ``{PATTY:} {A} taxonomy of
  relational patterns with semantic types,'' in \emph{EMNLP-CoNLL}.\hskip 1em
  plus 0.5em minus 0.4em\relax {ACL}, 2012, pp. 1135--1145.

\bibitem{MitchellCHTBCMG15}
T.~M. Mitchell, W.~W. Cohen, E.~R.~H. Jr., P.~P. Talukdar, J.~Betteridge,
  A.~Carlson, B.~D. Mishra, M.~Gardner, B.~Kisiel, J.~Krishnamurthy, N.~Lao,
  K.~Mazaitis, T.~Mohamed, N.~Nakashole, E.~A. Platanios, A.~Ritter, M.~Samadi,
  B.~Settles, R.~C. Wang, D.~T. Wijaya, A.~Gupta, X.~Chen, A.~Saparov,
  M.~Greaves, and J.~Welling, ``Never-ending learning,'' in
  \emph{{AAAI}}.\hskip 1em plus 0.5em minus 0.4em\relax {AAAI} Press, 2015, pp.
  2302--2310.

\bibitem{Zhang2004}
Z.~Zhang, ``Weakly-supervised relation classification for information
  extraction,'' in \emph{{CIKM}}.\hskip 1em plus 0.5em minus 0.4em\relax {ACM},
  2004, pp. 581--588.

\bibitem{MintzBSJ2009}
M.~Mintz, S.~Bills, R.~Snow, and D.~Jurafsky, ``Distant supervision for
  relation extraction without labeled data,'' in \emph{{ACL/IJCNLP}}.\hskip 1em
  plus 0.5em minus 0.4em\relax The Association for Computer Linguistics, 2009,
  pp. 1003--1011.

\bibitem{HoffmannZLZW2011}
R.~Hoffmann, C.~Zhang, X.~Ling, L.~S. Zettlemoyer, and D.~S. Weld,
  ``Knowledge-based weak supervision for information extraction of overlapping
  relations,'' in \emph{{ACL}}.\hskip 1em plus 0.5em minus 0.4em\relax The
  Association for Computer Linguistics, 2011, pp. 541--550.

\bibitem{Fellbaum1998}
C.~Fellbaum, Ed., \emph{{WordNet: an electronic lexical database}}, 1998.

\bibitem{NavigliP2012}
R.~Navigli and S.~P. Ponzetto, ``{B}abel{N}et: {T}he automatic construction,
  evaluation and application of a wide-coverage multilingual semantic
  network,'' \emph{Artificial Intelligence}, vol. 193, pp. 217--250, 2012.

\bibitem{CulottaS04}
A.~Culotta and J.~S. Sorensen, ``Dependency tree kernels for relation
  extraction,'' in \emph{{ACL}}, 2004, pp. 423--429.

\bibitem{Stevenson2005}
M.~Stevenson and M.~Greenwood, ``A semantic approach to {IE} pattern
  induction,'' in \emph{Proceedings of the 43rd Annual Meeting of the
  Association for Computational Linguistics (ACL'05)}.\hskip 1em plus 0.5em
  minus 0.4em\relax Association for Computational Linguistics, 2005, pp.
  379--386.

\bibitem{ZhouZ07}
G.~Zhou and M.~Zhang, ``Extracting relation information from text documents by
  exploring various types of knowledge,'' \emph{Inf. Process. Manage.},
  vol.~43, no.~4, pp. 969--982, 2007.

\bibitem{Moro13}
A.~Moro, H.~Li, S.~Krause, F.~Xu, R.~Navigli, and H.~Uszkoreit, ``Semantic rule
  filtering for web-scale relation extraction,'' in \emph{Proc. of ISWC}, 2013,
  pp. 347--362.

\bibitem{Nguyen2010}
Q.~L. Nguyen, D.~Tikk, and U.~Leser, ``Simple tricks for improving
  pattern-based information extraction from the biomedical literature,''
  \emph{Journal of Biomedical Semantics}, vol.~1, no.~1, pp. 1--17, 2010.

\bibitem{XuKPS02}
F.~Xu, D.~Kurz, J.~Piskorski, and S.~Schmeier, ``A domain adaptive approach to
  automatic acquisition of domain relevant terms and their relations with
  bootstrapping,'' in \emph{{LREC}}.\hskip 1em plus 0.5em minus 0.4em\relax
  European Language Resources Association, 2002.

\bibitem{Navigli2009}
R.~Navigli, ``Word sense disambiguation: A survey,'' \emph{ACM Comput. Surv.},
  vol.~41, no.~2, pp. 10:1--10:69, 2009.

\bibitem{Brin98}
S.~Brin and L.~Page, ``Anatomy of a large-scale hypertextual web search
  engine,'' in \emph{Proc. of WWW}, 1998, pp. 107--117.

\bibitem{Leskovec14}
J.~Leskovec, A.~Rajaraman, and J.~D. Ullman, \emph{Mining of Massive Datasets
  2nd Edition}.\hskip 1em plus 0.5em minus 0.4em\relax Cambridge University
  Press, 2014.

\bibitem{Tong06}
H.~Tong, C.~Faloutsos, and J.~Y. Pan, ``Fast random walk with restart and its
  applications,'' in \emph{Proc. of ICDM}, 2006.

\bibitem{Toutanova2015EMNLP}
\BIBentryALTinterwordspacing
K.~Toutanova, D.~Chen, P.~Pantel, H.~Poon, P.~Choudhury, and M.~Gamon,
  ``Representing text for joint embedding of text and knowledge bases,'' in
  \emph{Proceedings of the 2015 Conference on Empirical Methods in Natural
  Language Processing}.\hskip 1em plus 0.5em minus 0.4em\relax Lisbon,
  Portugal: Association for Computational Linguistics, September 2015, pp.
  1499--1509. [Online]. Available: \url{http://aclweb.org/anthology/D15-1174}
\BIBentrySTDinterwordspacing

\bibitem{Backstrom11}
L.~Backstrom and J.~Leskovec, ``Supervised random walks: Predicting and
  recommending links in social networks,'' in \emph{Proc. of WSDM}, 2011.

\end{thebibliography}
\bibliographystyle{IEEEtran}
%\normalsize


%\begin{IEEEbiography}[{\includegraphics[width=1in,height=1.25in,clip,keepaspectratio]{figures/alex}}]{Alexander Bauer}
%is a researcher at Berlin Big Data Center, Machine Learning Group, Technische Universit\"at Berlin.
%He received a B.Sc. in mathematics 2011 and a diploma in computer science in 2012, both from the Technische Universit\"{a}t Berlin, Germany. He is currently pursuing the Doctoral degree in the machine learning group of the Technische Universit\"{a}t Berlin. His scientific interests include theory and applications of machine learning, in particular, learning with structured outputs.
%\end{IEEEbiography}

\begin{IEEEbiography}[{\includegraphics[width=1in,height=1.25in,clip,keepaspectratio]{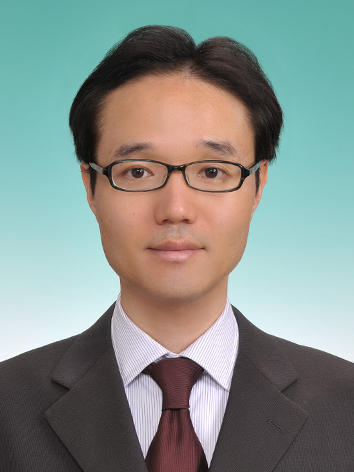}}]{Shinichi Nakajima}
is a senior researcher in Berlin Big Data Center, Machine Learning Group, Technische Universit\"at Berlin.  He received the master degree on physics in 1995 from Kobe university, and worked with Nikon Corporation until September 2014 on statistical analysis, image processing, and machine learning.  He received the doctoral degree on computer science in 2006 from Tokyo Institute of Technology.  His research interest is in theory and applications of machine learning, in particular, Bayesian learning theory, computer vision, and data mining.
\end{IEEEbiography}

\begin{IEEEbiography}[{\includegraphics[width=1in,height=1.25in,clip,keepaspectratio]{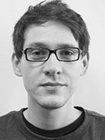}}]{Sebastian Krause}
%\begin{IEEEbiography}{Sebastian Krause}
is a PhD student at the Language Technology Lab of the German Research Center for Artificial Intelligence (DFKI). He got his Diplom degree in Computer Science from the Humboldt University Berlin and has recently worked on natural language processing, in particular on text mining problems.
\end{IEEEbiography}

\begin{IEEEbiography}[{\includegraphics[width=1in,height=1.25in,clip,keepaspectratio]{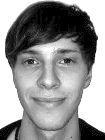}}]{Dirk Weissenborn}
%\begin{IEEEbiography}{Dirk Weissenborn}
is since April 2014 a Researcher und PhD Student at DFKI. His background is in Machine Learning with a focus on NLP.
\end{IEEEbiography}

\begin{IEEEbiography}[{\includegraphics[width=1in,height=1.25in,clip,keepaspectratio]{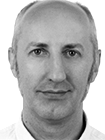}}]{Sven Schmeier}
%\begin{IEEEbiography}{Sven Schmeier}

is a senior consultant and project leader at the Language Technology Lab at the German Research Center for Artificial Intelligence (DFKI) in Berlin. Sven Schmeier holds a Diploma in Computer Science and a PhD in Computational Linguistics from the University of Saarland. In 2000 he was co-founder of the DFKI Spin-Off company Xtramind (now Attensity). In 2005 he was the leader of the research group at Semgine GmbH now reformed to medx GmbH in Berlin. In 2007 he was co-founder of the company Yocoy Technologies GmbH with Dr. Feiyu Xu and Prof. Hans Uszkoreit.
\end{IEEEbiography}

\begin{IEEEbiography}[{\includegraphics[width=1in,height=1.25in,clip,keepaspectratio]{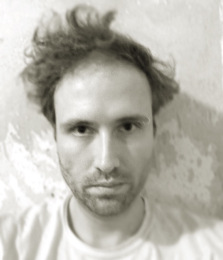}}]{Nico G{\"o}rnitz}
%\begin{IEEEbiography}{Nico G{\"o}rnitz}
is a research associate in the machine learning group at the Berlin Institute of Technology (TU Berlin, Berlin, Germany) headed by Klaus-Robert M\"uller. In 2014 he did an internship with the eScience Group, led by David Heckerman (Microsoft Research, Los Angeles, US). Before, he was employed as a research associate from 2010-2014 and during 2010-2012 also affiliated with the Friedrich Miescher Laboratory of the Max Planck Society in T\"ibingen, where he was co-advised by Gunnar R\"atsch. He received a diploma degree (MSc equivalent) in computer engineering (Technische Informatik) from the Berlin Institute of Technology with a thesis in machine learning for computer security in 2010.
\end{IEEEbiography}

%\begin{IEEEbiography}[{\includegraphics[width=1in,height=1.25in,clip,keepaspectratio]{klaus}}]{Klaus-Robert M\"uller}
%studied physics at University of Karlsruhe,
%Germany, from 1984 to 1989 and received the
%Ph.D. degree in computer science from University of Karlsruhe in 1992.
%He has been a Professor of computer science at
%Technische Universit{\"a}t Berlin, Berlin, Germany,
%since 2006. At the same time he has been the
%Director of the Bernstein Focus on Neurotechnology
%Berlin until 2013; from 2014 he has been Co-director of the Berlin Big Data Center. 
%After completing a postdoctoral position
%at GMD FIRST in Berlin, he was a Research Fellow at the University of
%Tokyo from 1994 to 1995. In 1995, he founded the Intelligent Data
%Analysis group at GMD-FIRST (later Fraunhofer FIRST) and directed it
%until 2008. From 1999 to 2006, he was a Professor at the University of
%Potsdam.
%Dr. M{\"u}ller was awarded the 1999 Olympus Prize by the German
%Pattern Recognition Society, DAGM, and, in 2006, he received the SEL
%Alcatel Communication Award. In 2014 he received the Berliner
%Wissenschaftspreis des regierenden B{\"u}germeisters. In 2012, he was
%elected to be a member of the German National Academy of Sciences-
%Leopoldina. His research interests are intelligent data analysis, machine
%learning, signal processing, and brain computer interfaces.
%\end{IEEEbiography}

\begin{IEEEbiography}[{\includegraphics[width=1in,height=1.25in,clip,keepaspectratio]{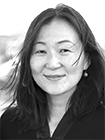}}]{Feiyu Xu}
%\begin{IEEEbiography}{Feiyu Xu}
 is Principal Researcher and Head of Research Group Text Analytics in the Language Technology Lab of DFKI. 
She also is co-founder of Yocoy Technologies GmbH, a 2007 spin-off from DFKI. Yocoy is developing next generation mobile language and travel guides.
Since 2004, Dr. Xu is vice-director of the Joint Research Laboratory for Language Technology of Shanghai Jiao Tong University and Saarland University.
Feiyu Xu studied technical translation at Tongji University in Shanghai after having been nominated and selected with a waiver of the national admission exam in year 1987. She then studied computational linguistics at Saarland University from 1992 to 1998 and graduated by receiving a Diplom (MSc) with distinction. Her PhD-Thesis is about "bootstrapping relation extraction from semantic seed" in "information extraction". In 2014, Feiyu Xu has completed a habilitation in big text data analytics.
In 2012, Feiyu Xu has won a Google Focused Research Award for Natural Language Understanding as co-PI with Hans Uszkoreit and Roberto Navigli.
In 2014, Feiyu Xu was honored as DFKI Research Fellow.
She has extensive experience in multilingual information systems, information extraction, text mining, big data analytics, business intelligence, question answering and mobile applications of NLP technologies. She has successfully led more than 30 national and international research and development projects. She has broad and in-depth experience of the total cycle of innovation in her expert areas, from basic research, to application and development and finally to products and their commercialization. 
\end{IEEEbiography}


%\vfill


\end{document}